\documentclass[12pt,english]{article}
\usepackage[cp866]{inputenc}
\usepackage[english]{babel}
\usepackage{amsmath}
\usepackage{graphicx}
\begin{document}
\begin{center}
{\Large\bf Factor Temporal Prognosis \\
of Tick-Borne Encephalitis Foci Functioning  on the South of
Russian Far East}
\medskip

{\bf E.I. Bolotin $\;\;\;\;\;\;\;\;\;\;\;\;\;\;\;\;\;\;\;\;\;\;\;\;\;\;\;\;\;\;\;\;\;\;\;\;\;\;\;\;\;\;\;\;\;\;\;\;$G.Sh. Tsitsiashvili\\
Pacific Ocean Institute of Geography $\;\;\;\;\;\;\;\;\;$ Institute of Applied Mathematics\\
  Far Eastern
Branch of RAS $\;\;\;\;\;\;\;\;\;\;\;\;\;\;\;\;\;\;$ Russia Far Eastern Branch of RAS\\
Vladivostok, Russia $\;\;\;\;\;\;\;\;\;\;\;\;\;\;\;\;\;\;\;\;\;\;\;\;\;\;\;\;\;\;\;\;\;\;\;$ Vladivostok, Russia\\
.$\;\;\;\;\;\;\;\;\;\;\;\;\;\;\;\;\;\;\;\;\;\;\;\;\;\;\;\;\;\;\;\;\;\;\;\;\;\;
\;\;\;\;\;\;\;\;\;\;\;\;\;\;\;\;\;\;\;\;\;\;\;\;\;\;\;\;\;\;\;\;\;\;\;\;\;\;\;\;\;\;\;$e-mail:guram@iam.dvo.ru\\
I.V. Golycheva \\
Institute of Applied Mathematics\\
Far Eastern Branch of RAS Vladivostok Russia\\
e-mail: iragol@yandex.ru }\end{center}
\medskip
\begin{center}
{\large\bf Abstract}
\end{center}

A method of temporal factor prognosis of TE (tick-borne
encephalitis) infection has been developed. The high precision of
the prognosis results for a number of geographical regions of
Primorsky Krai has been achieved. The method can be applied not
only to epidemiological research but also to others.\\\\

{\large\bf 1. Introduction}

While studying various complex systems the examination of their
functioning both in the past and in the future is of great
importance. The most informative index of natural pest foci is the
changing of human infection rate. Epidemic temporal rows of
infection integral expression of influence of a number of factors.
They comprise prognosis data which can be used to study this or
that phenomenon in future [1].

Prognosis approach in scientific research, including the
development of methodological base of natural - foci infection
rate prognosis, is the main research and practical goal. It is of
great importance nowadays because the epidemic situation has
become worse not only in Russia, but also in many regions and
countries of the world.

There exist two ways of temporal prognosis. The first one is
extrapolation prognosis. It is based on the analysis of temporal
rows of infection for the past several years and on their direct
extrapolation.

The second approach to temporal prognosis is a factor one and it
has several stages. Possibilities and advantages of temporal
factor prognosis are the following.

$\bullet$ First,  a researcher,  using extrapolation prognosis,
deals with average characteristics which reflect general   trend
of phenomenon development.  This trend is an artificial line where
the most important for prognosis of critical level conditions of
the analysed system are graded.

$\bullet$ Second, the results of the extrapolation prognosis
depend on the length of temporal raw and are specified by the type
of long standing development of the researched phenomenon.

$\bullet$ Third, the results of extrapolation prognosis have
probable character and are limited by confidence intervals. The
range of such limits, however, can be so broad that it makes the
obtained results so "vague" that considerably limits possibility
of their objective interpretation.

$\bullet$ Forth, extrapolation prognosis is realized by the "black
box" principle, it is deprived of the possibility to analyse the
caused factor and mechanism which form the temporal raw.

These limitations define our interest to more sophisticated and
yet more objective factor method in the prognosis of the epidemic
activity of tick-borne encephalitis, the area of which extends
from Europe to Japan [2], [3].

The levels of connection turned out to be unstable according to
what we have found out from our special preliminary research in
study of connection stability between the long standing row of
infection rate and some influential factors. Such unstable
correlation between infection rate and influential factors
strictly limits the possibilities of linear statistic models use.
Thus, the prognosis goal setting was changed. The nature of which
was to forecast indefinite index but infection rates which could
be higher or equal to some extreme line given by an expert. Such
goal setting is particularly important methodologically because
the way of its realisation has a unique character and can be
easily reproduced in other scientific spheres.

The factor prognosis research which we use has been divided into
two stages. The first stage included the realisation of the
epidemiological index and the questions concerning the potential
prognosis results accuracy by the different statistic selection
were also discussed. The aim of the second research stage was
firsthand factor prognosis of critical infection rate of tick-bome
encephalitis in a real time. Thus, while at the first stage of the
investigation purely academic problems were solved and
methodological approaches of putting it into practice were
developed, the second stage can be of practical importance not
only for tick-bome encephalitis but for wide range of other
phenomena as well.\\\\

{\large\bf 2. Main results of research}

The experimental material for factor temporal prognosis presents
statistic data on tick-bome infection dynamics during the last 30
years in a number of epidemic areas of Primorski Krai, previously
determined by us [4].

On prognosis on the first investigational stage the use was made
of influential factors that are long standing rows of average
annual meaning of 8 meteorological factors of 3 meteorological
stations, that characterise southern, middle and northern regions
of Primorski Krai. The used meteorological factors include:
average January temperature, absolute temperature minimum, average
annual temperature, the length of non-frosty period, average May
temperature, average snow blanket and snow-temperature
coefficient.

On the second stage of temporal factor prognosis the materials on
long standing dynamics of population Ixodes persulcatus - the main
TE virus carrier were drawn in addition.

Methods of factor prognosis of critical levels of TE infection
were based on the following algorithm worked out by us. Firstly,
empirical information of TE and influential factors was introduced
as the matrix of initial data in which rows indicate years and
columns indicate index of infection and influential factors.
Further, the years rows with critical level of infection and
corresponding indexes of influential factors were emphasized. The
emphasized years with critical levels of infection form intervals
of particular influential factor meanings. However, these
intervals may also include years in which the infection is lower
than critical level that we emphasized earlier. Let's mark such
years as "false critical". Having marked the amount of critical
and "false critical" years as "$x$" and "$y$" accordingly. We have
a ratio $p=x/(x+y)$ which can be interpreted as possibility to
identify correctly the critical years on emphasized intervals of
influential factors.

As a whole the developed method of factor temporal prognosis was
based on the image recognition principle. But the quality
prognosis, realized by original algorithm, was determined
according to selection by number of non-critical years falsie
taken for critical by the determined final rule of recognition.
And at the first stage of investigation the final rule of the year
recognition as a critical was based on the condition of 100$\%$
this year being in the defined meaning intervals of influential
factors of critical years.

However, at the second stage of investigation i.e. on prognosis of
critical events in real time, it turned out that such condition is
just of secondary meaning. This conclusion was made during
numerous calculating experiments that we conducted.

As a result, new prognosis rule was established in which a number
of identifying factors fluctuated between 50 and 75 percents of
all the used ones. Critical infection rate was also chosen
depending on statistic sample with the amount of critical years
being more that one. It should be noted that during the repeated
besting of prognosis rule, simplicity and speed of included
calculations were of the primary importance.

On the whole, some most significant moments of conducted factor
prognosis can be signed out. Thus, accuracy of factor prognosis
totalled from 50 to 100 percents.

The maximum prognosis rate was provided southerner territories of
investigated region, however, some other geographic regions
characterized by close prognosis figures were mentioned.

The prognosis rate is considerably influenced by the amount and
concrete set of affecting factors while the increase of used
factors number does not always result in higher prognosis
accuracy. This remarkable fact requires further thorough
investigation and specification.

The received data concerning the connection between prognosis
accuracy and used critical infection rate are also interesting. It
was found out that in some cases this connection can be direct and
indirect in the others. This quite substantial fact also requires
serious analysis and consideration.

It was revealed that the usage of factual materials on long-term
Ixodes perculcatus's population dynamics had not improved
prognosis accuracy. It remained the same or even lowered. This
tendency again proves our opinion about the insignificant
influence of carriers population on human's suffering from
tick-borne Encephalitis. If we compare the results of both the
first and second stages of temporary factor prognosis, their
considerable similarity can be marked.\\\\

{\large\bf 3. Conclusion}

On the whole, the developed method of critical infection rates
temporary factor prognosis solves an extremely vital problem of
non-linear .character, from the other side it has several
important features.

$\bullet$ First, this method is quite easy for realization, but at
the same time it is all-purpose; namely it is able to work with
any information presented in the form of dynamic rows.

$\bullet$ Second, all the calculations performed with the aid of
prepared program can easily be conducted manually.

$\bullet$ Third, the proposed method can be regarded as one of the
highest level steps for real prognosis , based on dynamic
cause-and-effect models.

$\bullet$ Forth, the method of critical disease rate analysis has
good perspectives  as it deals with  a  number of important
problems. For  example, how  will the results of the prognosis
change if the number and nature of the factors are altered? What
is the effect of modifying the length of rows and critical disease
ranges? How can the quality of the program of the prognosis be
modified  in case  of correlating the disease and the influential
factors with different logs? How is the influential factors
hierarchy modified with regard to their prognosis importance in
the course of temporal rows analysis? There also exits quite a
number of vitally important problems. This work is the first and
quite a modest attempt to solve some of these problems.

Thus, the developed temporal factor prognosis method being of both
theoretical and practical interest, poses a number of important
questions as well. Having solved the problems by means of
wide-range testing of the proposed algorithm, it is possible to
receive, for the future, a universal method of temporal factor
prognosis.\\\\

{\large\bf References}\\
1. Iagodinsky V.N. "Dynamics of Epidemiological Process", Moscow,
Medicine, 1977, 240 p.\\
2. Zlobin V.I., Gorin O.Z. "Tick-Borne Encephalitis", Novosibirsk,
Medicine, 1996, 177 p.\\
3. Korenberg A.I. "Biohorological Structure of Species (on Example
of Taiga Tick)", Moscow, Science, 1979, 172 p.\\
4. Bolotin E.I. "On a Functional Organization of Natural Tick-Bome
Encephalitis Foci and Prognosis of Their Epidemic Manifestation:
Analysis of Monodimensional Temporal Rows of Infection".
Parazitology, 2001; 35, 5, 386-393.

\end{document}